\documentclass[10pt,conference]{IEEEtran}

\newif\ifieee
\ieeefalse

\newif\ifarxiv
\arxivtrue

\ifieee
    \ifarxiv
        \PackageError{Configuration}
        {IEEE and arXiv cannot both be enabled}
        {Set either \string\ieeefalse\space or \string\arxivfalse.}
    \fi
\fi

\usepackage{multirow} %
\usepackage{makecell}

\usepackage{arydshln}
\usepackage[normalem]{ulem}
\usepackage{balance}
\usepackage{newtxtext} %
\usepackage{xspace}

\usepackage[sort,compress]{cite}

\ifCLASSINFOpdf
   \usepackage[pdftex]{graphicx}
   \graphicspath{{figs/}}
   \DeclareGraphicsExtensions{.pdf,.jpeg,.png}
\else
   \usepackage[dvips]{graphicx}
   \graphicspath{{../figs/}}
   \DeclareGraphicsExtensions{.eps}
\fi

\usepackage[cmex10]{amsmath}
\usepackage{textcomp} %
\usepackage{gensymb} %
\usepackage{amsthm}

\usepackage{algorithmic}

\usepackage{array}

\ifCLASSOPTIONcompsoc
  \usepackage[caption=false,font=normalsize,labelfont=sf,textfont=sf,farskip=0pt]{subfig}
\else
  \usepackage[caption=false,font=footnotesize,farskip=0pt]{subfig}
\fi

\usepackage[acronym]{glossaries}
\glsdisablehyper
\usepackage{xspace}
\usepackage[hyphens]{url}
\usepackage{booktabs}
\usepackage[dvipsnames]{xcolor}%
\usepackage[utf8]{inputenc}
\usepackage[T1]{fontenc}
\usepackage[colorlinks=true,allcolors=black]{hyperref} %

\usepackage[capitalise]{cleveref} %

\newacronymstyle{long-short-br}
{%
  \GlsUseAcrEntryDispStyle{long-short}%
}%
{%
  \GlsUseAcrStyleDefs{long-short}%
}
\setacronymstyle{long-short-br}

\usepackage{transparent}
\usepackage{tikz}
\ifarxiv
    \newcommand\copyrighttext{%
      \scriptsize Accepted for presentation at SIBGRAPI 2026. The final published version will be available on IEEE~Xplore.}
    \newcommand\copyrightnotice{%
    \begin{tikzpicture}[remember picture,overlay]
    \node[anchor=south,yshift=30pt,xshift=0pt] at (current page.south) {\fbox{\transparent{0.85}\parbox{\dimexpr0.6\textwidth-\fboxsep-\fboxrule\relax}{\copyrighttext}}};
    \end{tikzpicture}%
    }
\else
\fi

\newif\iffinal
\finaltrue
\newcommand{\cmtid}{107}

\iffinal
\else
    \usepackage[switch]{lineno}
\fi

\newcommand*{\RL}[2][]{\textcolor{Rhodamine}{[\textbf{\ifthenelse{\equal{#1}{}}{RL}{RL(#1)}}: #2]}}
\newcommand*{\DM}[2][]{\textcolor{orange}{[\textbf{\ifthenelse{\equal{#1}{}}{DM}{DM(#1)}}: #2]}}
\newcommand*{\LW}[2][]{\textcolor{cyan}{[\textbf{\ifthenelse{\equal{#1}{}}{LW}{LW(#1)}}: #2]}}
\newcommand*{\GL}[2][]{\textcolor{brown}{[\textbf{\ifthenelse{\equal{#1}{}}{GL}{GL(#1)}}: #2]}}
\newcommand*{\SJ}[2][]{\textcolor{red}{[\textbf{\ifthenelse{\equal{#1}{}}{SJ}{SJ(#1)}}: #2]}}
\newcommand*{\OR}[2][]{\textcolor{olive}{[\textbf{\ifthenelse{\equal{#1}{}}{OR}{OR(#1)}}: #2]}}

\newcommand\major[1]{#1} %

\newcounter{fncounter}
\ifieee
\IEEEoverridecommandlockouts
\IEEEpubid{\makebox[\columnwidth]{979-8-3195-0255-1/26/\$31.00~\copyright2026 IEEE \hfill}
\hspace{\columnsep}\makebox[\columnwidth]{ }}
\else
\fi

\begin{document}

\iffinal
    \newcommand{\urlSupplementary}{\url{https://github.com/UFPR-IPASP-PR/uvib-vehicle-attributes/}}
\fi

\newcommand{\benchmarkName}{UVIB\xspace}
\newcommand{\numImages}{$84{,}835$\xspace}

\title{A Benchmark for Vehicle Attribute Classification\\in Cross-Domain Surveillance Scenarios} %

\iffinal

\author{
\IEEEauthorblockN{Sergio M. {Silva~Jr.}\IEEEauthorrefmark{1}, Otavio T. Remer\IEEEauthorrefmark{1}, Gabriel E. Lima\IEEEauthorrefmark{1},\\Lucas Wojcik\IEEEauthorrefmark{1}, Rayson~Laroca\IEEEauthorrefmark{2}$^,$\IEEEauthorrefmark{1}, and David~Menotti\IEEEauthorrefmark{1}\\[0.75ex]}
\IEEEauthorblockA{
    \IEEEauthorrefmark{1}Department of Informatics, Federal University of Paran\'{a}, Curitiba, Brazil \\    \IEEEauthorrefmark{2}\hspace{0.1mm}Graduate
    Program in Informatics, Pontifical Catholic University of Paran\'a, Curitiba, Brazil \\[0.75ex]
        \hspace{-0.75mm}\IEEEauthorrefmark{1}\hspace{-0.35mm}\tt{\small{\{smsjunior,otavio,gelima,lmlwojcik,menotti\}}@inf.ufpr.br} \ \quad \IEEEauthorrefmark{2}{\tt\small rayson@ppgia.pucpr.br}}
}

\fi

\maketitle

\ifarxiv
    \copyrightnotice
\else
\fi

\newacronym{alpr}{ALPR}{Automatic License Plate Recognition}
\newacronym{c2psa}{C2PSA}{Cross Stage Partial with Spatial Attention}
\newacronym{capes}{CAPES}{Coordination for the Improvement of Higher Education Personnel}
\newacronym{cbs}{CBS}{Convolution-BatchNorm-SiLU}
\newacronym{cnn}{CNN}{Convolutional Neural Network}
\newacronym{cnpq}{CNPq}{National Council for Scientific and Technological Development}
\newacronym{f1}{F1}{F1-score}
\newacronym{fgvc}{FGVC}{Fine-Grained Vehicle Classification}
\newacronym{hc-err}{HC-Err}{Hierarchical Consistency Error}
\newacronym{hfgvc}{HFGVC}{Hierarchical Fine-Grained Visual Classification}
\newacronym{its}{ITS}{Intelligent Transportation Systems}
\newacronym{kl}{KL}{Kullback–Leibler}
\newacronym{ma-acc}{Ma-Acc}{Macro-Accuracy}
\newacronym{mi-acc}{Mi-Acc}{Micro-Accuracy}
\newacronym{mtl}{MTL}{Multitask Learning}
\newacronym{senatran}{SENATRAN}{Brazil's National Traffic Secretariat}
\newacronym{uvib}{UVIB}{Unconstrained Vehicle Identification Benchmark}
\newacronym{vcr}{VCR}{Vehicle Color Recognition}
\newacronym{vit}{ViT}{Vision Transformer}
\newacronym{vmmr}{VMMR}{Vehicle Make and Model Recognition}
\newacronym{reid}{ReID}{re-identification}

\ifieee
\vspace{-3.575mm}
\else
\vspace{-3.575mm}
\fi

\begin{abstract}
Vehicle attribute analysis is a key component of \gls*{its}, supporting applications such as vehicle identification, traffic monitoring, and forensic investigation. However, models trained under controlled conditions often degrade in real surveillance scenarios due to changes in viewpoint, occlusion, illumination, and sensor characteristics. This paper introduces \gls*{uvib}, a benchmark for evaluating three operational vehicle-analysis tasks: front/rear orientation, occlusion-related suitability for \gls*{vmmr}, and color clarity. The benchmark contains 84,835 vehicle images from seven public Brazilian datasets, grouped into surveillance and general acquisition domains, with unified binary annotations that were not jointly available in the original sources. Four representative architectures, EfficientNetV2-S, ResNet-50, ViT/B-16, and YOLO11s-cls, are evaluated under mixed-domain, cross-domain, and cross-dataset protocols. The results show that domain shift has a stronger impact than architecture choice, with substantial degradation in cross-domain settings, especially for \gls*{vmmr} suitability and color clarity. While orientation generalizes more reliably, \gls*{vmmr} suitability remains affected by class imbalance and ambiguous occlusions, and color clarity is highly sensitive to illumination and sensor modality. These findings highlight the need for benchmarks and evaluation protocols that explicitly measure operational robustness beyond standard in-domain accuracy.
\major{The proposed benchmark is publicly available at \textit{\urlSupplementary}}.
\end{abstract}

\IEEEpeerreviewmaketitle

\section{Introduction}
\label{sec:introduction}

\glsresetall

\gls*{its} enable the automatic extraction of vehicle attributes used in identification, traffic monitoring, and forensic analysis, including make, model, color, and license plate information~\cite{gayen2024two,alzahrani2024survey,kerdvibulvech2025multimodal,lima2026toward}.
However, models evaluated under controlled conditions often degrade in urban monitoring scenarios, where viewpoint, occlusion, illumination, and sensor characteristics vary substantially~\cite{gayen2024two,laroca2026competition}.
This work focuses on three operational decisions that affect downstream vehicle analysis: orientation, occlusion-related suitability for make and model recognition, and color~clarity.

Distinguishing frontal from rear views is important because logos, grilles, taillights, and license plates provide different evidence for accurate vehicle identification
depending on the viewpoint~\cite{oliveira2021vehicle,amirkhani2022deepcar,gayen2024two,lima2026toward}.
\gls*{vmmr} Suitability refers to whether a vehicle crop preserves enough visible structure for reliable make and model analysis; vehicles with key regions hidden by other vehicles, trees, or traffic infrastructure may be unsuitable even if they are still detectable~\cite{zhang2008multilevel,tian2015vehicle,amirkhani2022deepcar,gayen2024two,wojcik2025lplc}.
Color clarity indicates whether the image contains reliable chromatic evidence for vehicle color analysis. A non-color sample is not a vehicle without a physical color; rather, it is an image in which color cannot be confidently inferred because of infrared capture, severe illumination, sensor degradation, or similar factors~\cite{hsieh2015vehicle,lima2024toward,orru2026revisiting}.

\begin{figure}[!htbp]
\centering
    \begin{minipage}{0.31\linewidth}
        \centering
        {\footnotesize (a) Orientation} \\[0.75mm] %
        \includegraphics[width=0.95\linewidth, height=2.3cm]{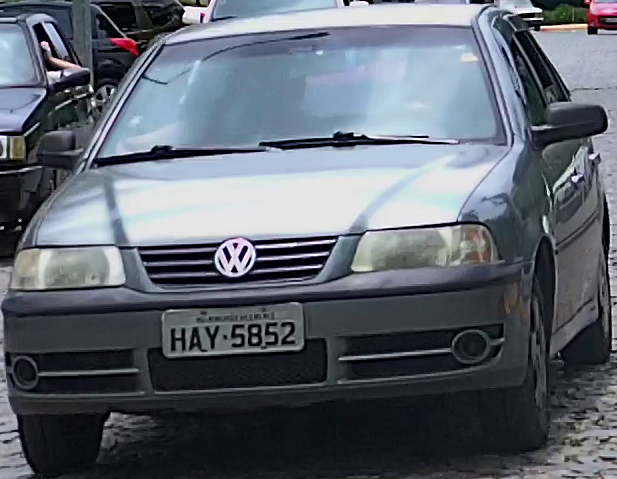} \\[-1.75mm] %
        {\scriptsize Front} \\[2.75mm] %
        \includegraphics[width=0.95\linewidth, height=2.3cm]{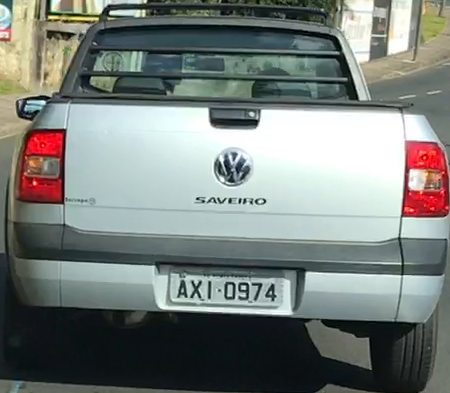} \\[-1.75mm]
        {\scriptsize Rear}
    \end{minipage}
    \hfill
    \begin{minipage}{0.31\linewidth}
        \centering
        {\footnotesize (b) \gls*{vmmr} Suitability} \\[0.75mm]
        \includegraphics[width=0.95\linewidth, height=2.3cm]{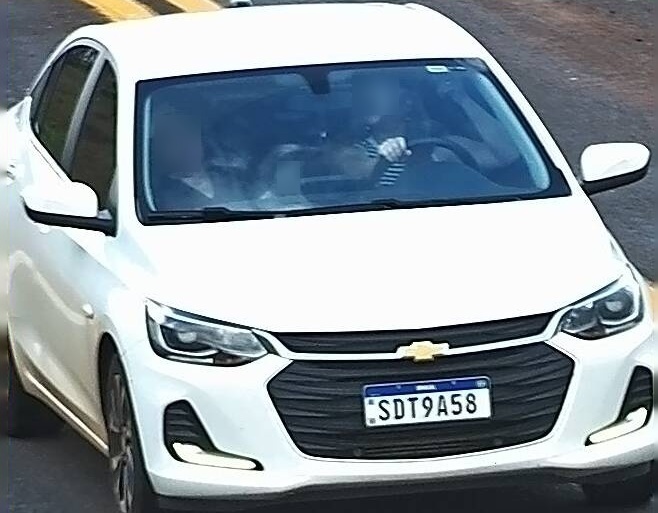} \\[-1.75mm]
        {\scriptsize Suitable} \\[2.75mm]
        \includegraphics[width=0.95\linewidth, height=2.3cm]{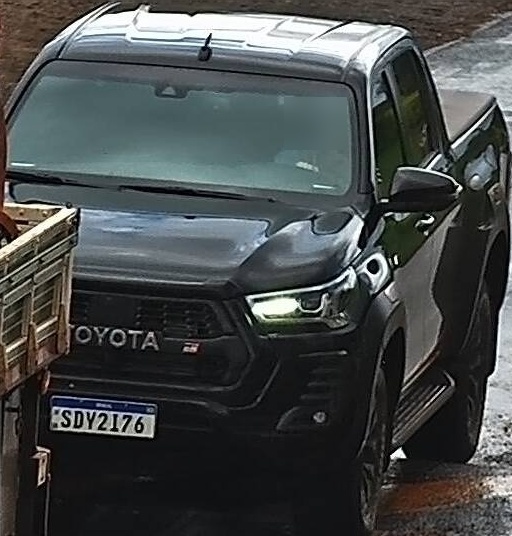} \\[-1.75mm]
        {\scriptsize Unsuitable}
    \end{minipage}
    \hfill
    \begin{minipage}{0.31\linewidth}
        \centering
        {\footnotesize (c) Color Clarity} \\[0.75mm]
        \includegraphics[width=0.95\linewidth, height=2.3cm]{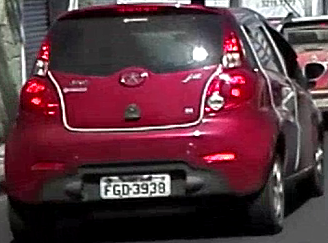} \\[-1.75mm]
        {\scriptsize Color} \\[2.75mm]
        \includegraphics[width=0.95\linewidth, height=2.3cm]{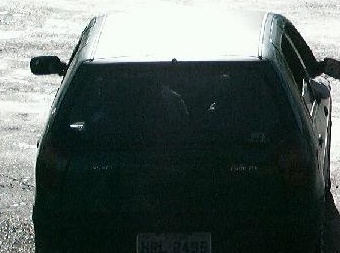} \\[-1.75mm]
        {\scriptsize Non-Color}
    \end{minipage}

\caption{Examples of the three target tasks in the proposed \acrfull*{uvib}.
Orientation separates front and rear views. \gls*{vmmr} Suitability indicates whether occlusion still allows reliable make and model analysis. Color Clarity distinguishes images with reliable chromatic evidence from non-color cases, where vehicle color cannot be confidently inferred from the image.
}

\label{fig:taxonomy_samples}
\end{figure}

Examples of these tasks are shown in \cref{fig:taxonomy_samples}. Existing vehicle datasets are usually designed for related goals, such as \gls*{vmmr}, \gls*{alpr}, or \gls*{vcr}, and therefore rarely provide unified labels for the three operational factors considered here. This limits direct comparison across datasets and makes it difficult to isolate the impact of dataset bias and domain shift on these decisions.

To address this gap, we introduce the \gls*{uvib}, composed of \numImages images from seven public Brazilian datasets~\cite{oliveira2021vehicle,wojcik2026lplcv2, lima2026toward, mendesjunior2011towards, goncalves2016benchmark, laroca2018robust, laroca2022cross}. We provide new labels for vehicle Orientation~(front/rear), \gls*{vmmr} Suitability~(suitable/unsuitable), and Color Clarity~(color/non-color), none of which were jointly available in the original sources. The benchmark combines surveillance and general-domain imagery, enabling controlled analyses of acquisition-domain~transfer.

This paper makes the following contributions:
\begin{itemize}
\item We create and release \gls*{uvib}, a unified benchmark with annotations for three operational vehicle-analysis tasks across seven public datasets;
\item We provide standardized baselines for four representative deep architectures under four evaluation protocols, including strict cross-domain settings;
\item We analyze how domain shift affects each task, showing that \gls*{vmmr} suitability and color clarity remain the main bottlenecks for robust deployment.
\end{itemize}

\section{Related Work}
\label{sec:related_work}

Vehicle attribute analysis supports several \gls*{its} applications, including \gls*{vmmr}, \gls*{vcr}, \gls*{alpr}, and vehicle \gls*{reid}~\cite{oliveira2021vehicle, nascimento2025toward, wojcik2025lplc, peng2020vehicle, gayen2024two}. These tasks rely on visual cues that may be unreliable in surveillance imagery due to viewpoint changes, occlusions, and color variations. This section reviews the attributes studied and the impact of dataset bias on cross-domain vehicle~analysis.

\subsection{Vehicle Attribute Recognition in Challenging Scenarios}

\gls*{vmmr} and related vehicle-recognition tasks have been studied with increasingly strong deep models and large-scale datasets~\cite{yang2015compcars,sochor2016boxcars,amirkhani2022deepcar,gayen2024two,kerdvibulvech2025multimodal,lima2026toward}. 
However, many of these methods assume that the input image already contains a suitable vehicle view. Orientation is therefore often used implicitly or as auxiliary context: frontal views emphasize grilles, headlights, and emblems, whereas rear views emphasize taillights, trunk regions, and license plates~\cite{oliveira2021vehicle,amirkhani2022deepcar,tan2025cross,lima2026toward}. When this information is not controlled, the same vehicle category may present different discriminative cues across datasets.

Occlusion handling has also been investigated in traffic monitoring, especially for overlapping vehicles and partial blockage caused by road infrastructure~\cite{zhang2008multilevel,tian2015vehicle,chang2018vision}. These works show that visible parts, such as headlights, taillights, and motion-consistent regions, are critical for separating or classifying vehicles. In attribute-recognition pipelines, the same issue appears as an operational question: whether the remaining visible structure is sufficient for a downstream task such as~\gls*{vmmr}.

Color recognition is affected by illumination, exposure, reflections, and sensor modality~\cite{chen2014vehicle,hsieh2015vehicle,lima2024toward}. Recent studies report that such factors can make color labels unreliable, even when the vehicle itself is visible~\cite{orru2026revisiting,wojcik2026lplcv2}.
This motivates color clarity as a prior decision that determines whether a color-recognition result should be trusted.

\subsection{Dataset Bias and Cross-Domain Generalization}

Dataset bias is a long-standing problem in computer vision: models may exploit acquisition-specific shortcuts rather than the visual concepts they are expected to learn~\cite{torralba2011unbiased}. In vehicle analysis, this issue is amplified by differences in camera placement, compression, illumination, and sensor technology~\cite{laroca2022first}.

Cross-dataset evaluations in \gls*{alpr}, \gls*{vmmr}, and \gls*{vcr} consistently report performance drops when training and testing data come from different sources~\cite{laroca2022cross,gayen2024two,lima2026toward}. These drops indicate that strong in-domain accuracy does not guarantee robustness under deployment shifts. Orientation, \gls*{vmmr} suitability, and color clarity are especially sensitive to this problem because the cues used for each decision depend directly on acquisition conditions.

The proposed benchmark complements previous vehicle datasets by providing unified annotations for these three tasks and pairing them with protocols that separate mixed-domain performance from cross-domain transfer. This design enables a more systematic analysis of which attributes generalize and which remain constrained by dataset-specific~characteristics.

\section{The \benchmarkName Benchmark}
\label{sec:benchmark}

\gls*{uvib} contains \numImages vehicle images aggregated from seven public Brazilian datasets. The benchmark is organized into two acquisition domains to support explicit domain-transfer analyses.  The \textit{Surveillance Domain} contains $57{,}798$ images from fixed traffic-monitoring cameras, while the \textit{General Domain} contains $27{,}037$ images collected under more heterogeneous viewpoints, environments, and acquisition setups.
This separation enables assessing whether models trained with one type of visual evidence remain reliable when evaluated under another (see the explored protocols in \cref{subsec:protocols}).

\begin{figure}[htbp]
\centering
    \resizebox{0.99\linewidth}{!}{
    \begin{minipage}{0.31\linewidth}
        \centering
        \includegraphics[width=\linewidth, height=2.3cm]{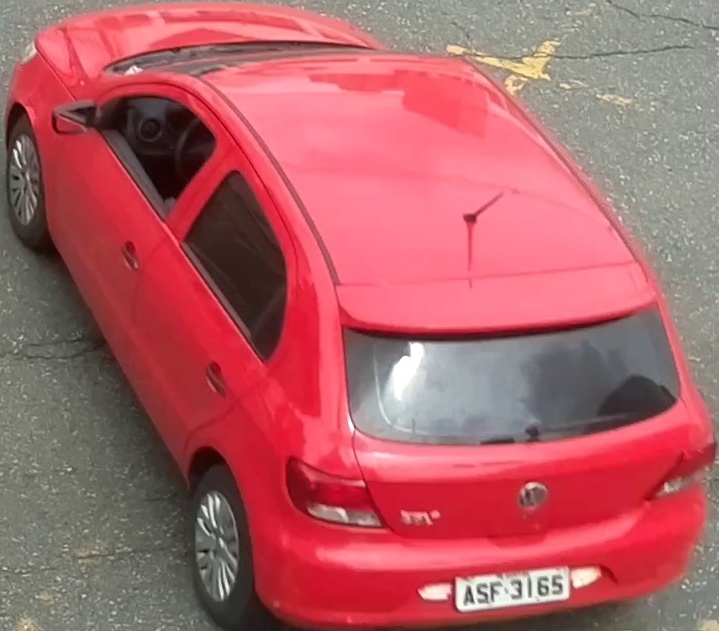} \\[-1.5mm]
        {\scriptsize Vehicle-Rear}
    \end{minipage}
    \hfill
    \begin{minipage}{0.31\linewidth}
        \centering
        \includegraphics[width=\linewidth, height=2.3cm]{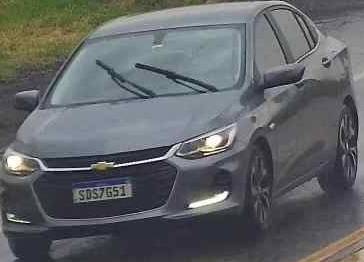} \\[-1.5mm]
        {\scriptsize LPLCv2}
    \end{minipage}
    \hfill
    \begin{minipage}{0.31\linewidth}
        \centering
        \includegraphics[width=\linewidth, height=2.3cm]{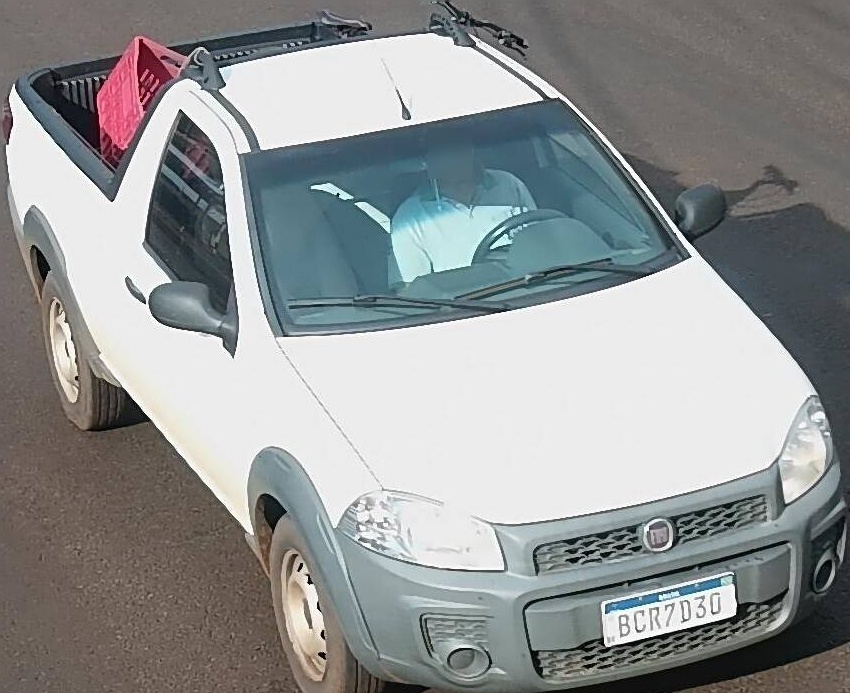} \\[-1.5mm] %
        {\scriptsize UFPR-VeSV}
    \end{minipage}
    }

    \vspace{1.25mm}

    {\footnotesize (a) Surveillance Domain} \\[0.5mm]

    \vspace{2mm}

    \resizebox{0.99\linewidth}{!}{
    \begin{minipage}{0.23\linewidth}
        \centering
        \includegraphics[width=\linewidth, height=1.8cm]{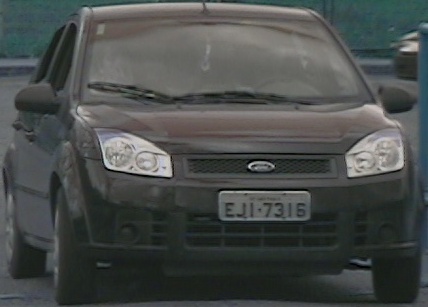} \\[-1.5mm]
        {\scriptsize UFOP}
    \end{minipage}
    \hfill
    \begin{minipage}{0.23\linewidth}
        \centering
        \includegraphics[width=\linewidth, height=1.8cm]{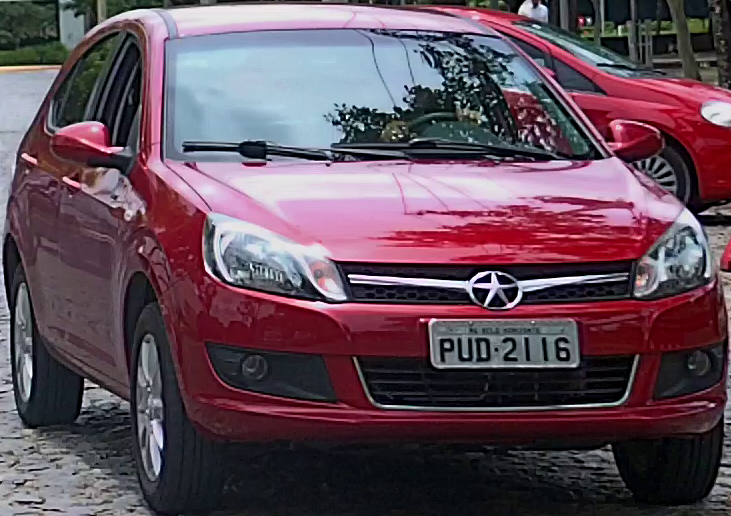} \\[-1.5mm]
        {\scriptsize SSIG-SegPlate}
    \end{minipage}
    \hfill
    \begin{minipage}{0.23\linewidth}
        \centering
        \includegraphics[width=\linewidth, height=1.8cm]{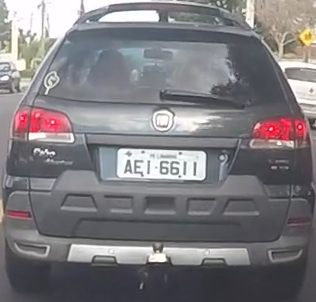} \\[-1.5mm]
        {\scriptsize UFPR-ALPR}
    \end{minipage}
    \hfill
    \begin{minipage}{0.23\linewidth}
        \centering
        \includegraphics[width=\linewidth, height=1.8cm]{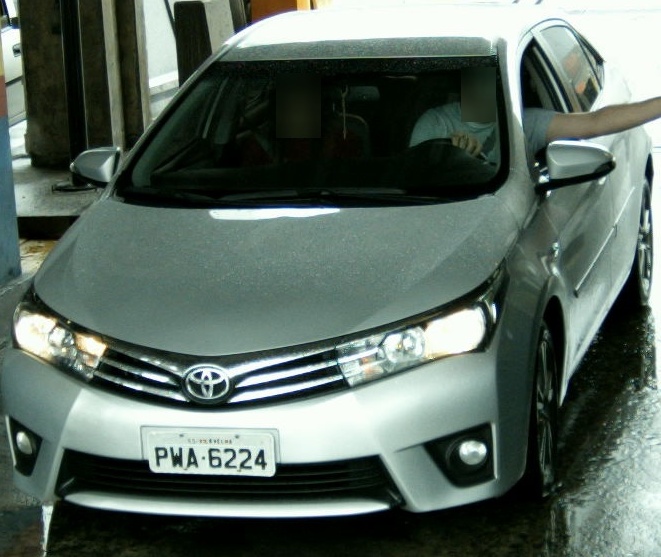} \\[-1.5mm]
        {\scriptsize RodoSol-ALPR}
    \end{minipage}
    }

    \vspace{1.25mm}
    
    {\footnotesize (b) General Domain}

\caption{Representative vehicle crops from the seven datasets used in \gls*{uvib}, grouped by acquisition domain. Datasets within each domain are shown in chronological publication~order.}
\label{fig:dataset_samples}
\end{figure}

The \textit{Surveillance Domain} includes Vehicle-Rear~\cite{oliveira2021vehicle}, LPLCv2~\cite{wojcik2026lplcv2}, and UFPR-VeSV~\cite{lima2026toward}.
The \textit{General Domain} includes UFOP~\cite{mendesjunior2011towards}, SSIG-SegPlate~\cite{goncalves2016benchmark}, UFPR-ALPR~\cite{laroca2018robust}, and RodoSol-ALPR~\cite{laroca2022cross}.
These datasets were selected because they are widely used in Brazilian vehicle and \gls*{alpr} research, cover diverse illumination and viewpoint conditions, and provide a realistic basis for studying dataset shift~\cite{laroca2022first,ismail2025automatic}. \cref{fig:dataset_samples} shows representative crops from each source.

The vehicle cropping process is detailed in \cref{sec:curation}, followed by the formal definitions and criteria established for our annotation framework in \cref{sec:annotation}.

\subsection{Preprocessing and Data Extraction}
\label{sec:curation}

Different extraction strategies were applied depending on the annotations available in each source dataset. UFPR-VeSV already provides cropped vehicle instances and therefore required no additional extraction. Vehicle-Rear, SSIG-SegPlate, and UFPR-ALPR provide vehicle bounding boxes, which were used directly. In contrast, LPLCv2, RodoSol-ALPR, and UFOP required an automated YOLO-based extraction pipeline.
As some source images contain multiple vehicles, the final number of vehicle crops may exceed the number of original~images.

\subsection{Annotation Framework}
\label{sec:annotation}

Each vehicle crop receives three independent binary annotations. Orientation indicates whether the vehicle is observed from the front or rear. \gls*{vmmr} Suitability captures whether the crop provides sufficient visual evidence for make and model analysis. Suitable images preserve enough visible frontal or rear structure for recognition, whereas unsuitable images exhibit severe occlusion or lack key regions such as headlights, taillights, the grille, or the trunk. Both Orientation and \gls*{vmmr} Suitability were manually annotated for every vehicle~sample.

Color clarity indicates whether chromatic information is visually reliable for downstream vehicle analysis. Images are labeled as non-color when color cannot be confidently inferred from the image because of poor illumination, severe sensor degradation, infrared capture, or other acquisition conditions. Thus, non-color is an image-quality and sensing label, not a vehicle-color category. 
Visible-spectrum images may also be non-color when chromatic cues are strongly degraded, as illustrated in \cref{fig:taxonomy_samples}.

The UFPR-VeSV dataset provides binary metadata indicating whether an image was captured by an infrared sensor. To distinguish infrared imagery from the broader non-color concept, two random UFPR-VeSV subsets, with $2{,}000$ infrared and $2{,}000$ non-infrared images, were manually annotated for color clarity by two independent annotators.
The same procedure was applied to $5{,}000$ random images from the remaining datasets. Inter-annotator agreement was measured using accuracy and Cohen's Kappa~\cite{viera2005understanding}, yielding Kappa values of $0.69$ for UFPR-VeSV and $0.62$ for the multi-dataset subset.
These values indicate substantial agreement, while also confirming that color clarity includes borderline cases that require explicit~auditing.

To scale the annotation process to the complete set of \numImages images, we adopted a multi-stage auditing protocol. Initial disagreements and borderline cases were isolated, reconciled manually, and reviewed in consecutive rounds. Finally, an exhaustive consistency sweep was conducted over the final dataset to reduce residual annotation noise.

The final dataset contains binary labels for orientation, \gls*{vmmr} suitability, and color clarity. Detailed distributions are reported in \cref{tab:dataset_composition}.

\begin{table*}[htb!]
    \caption{Sample distribution by source dataset, acquisition domain, and target annotation. Datasets are grouped by domain and ordered chronologically within each domain.}
    \label{tab:dataset_composition}

    \vspace{-0.5mm}
    
    \centering
    \begin{tabular}{lcccccccc}
        \toprule
        \multirow{2}[2]{*}{\textbf{Source Dataset}} & \multirow{2}[2]{*}{\textbf{Domain}} & \multicolumn{2}{c}{\textbf{Orientation}} & \multicolumn{2}{c}{\textbf{\gls*{vmmr} Suitability}} & \multicolumn{2}{c}{\textbf{Color Clarity}} & \multirow{2}[2]{*}{\textbf{Images}} \\ \cmidrule{3-8}
         &  & \textbf{Front} & \textbf{Rear} & \textbf{Suitable} & \textbf{Unsuitable} & \textbf{Color} & \textbf{Non-Color} &  \\ \midrule
        Vehicle-Rear~\cite{oliveira2021vehicle} & Surveillance & $\phantom{00{,}00}0$ & $\phantom{00{,}}730$ & $\phantom{00{,}}641$ & $\phantom{0{,}0}89$ & $\phantom{00{,}}390$ & $\phantom{00{,}}340$ & $\phantom{00{,}}730$ \\
        LPLCv2~\cite{wojcik2026lplcv2} & Surveillance & $\phantom{0}2{,}816$ & $29{,}307$ & $24{,}413$ & $7{,}710$ & $19{,}046$ & $13{,}077$ & $32{,}123$ \\
        UFPR-VeSV~\cite{lima2026toward} & Surveillance & $11{,}103$ & $13{,}842$ & $23{,}490$ & $1{,}455$ & $18{,}612$ & $\phantom{0}6{,}333$ & $24{,}945$ \\
        \midrule
        UFOP~\cite{mendesjunior2011towards} & General & $\phantom{0{,}0}298$ & $\phantom{00{,}}173$ & $\phantom{00{,}}362$ & $\phantom{0{,}0}109$ & $\phantom{00{,}}408$ & $\phantom{00{,}}63$ & $\phantom{00{,}}471$ \\
        SSIG-SegPlate~\cite{goncalves2016benchmark} & General & $\phantom{0}2{,}066$ & $\phantom{00{,}00}0$ & $\phantom{0}1{,}990$ & $\phantom{0{,}0}76$ & $\phantom{0}1{,}715$ & $\phantom{00{,}}351$ & $\phantom{0}2{,}066$ \\
        UFPR-ALPR~\cite{laroca2018robust} & General & $\phantom{00{,}00}0$ & $\phantom{0}4{,}500$ & $\phantom{0}4{,}392$ & $\phantom{0{,}}108$ & $\phantom{0}2{,}739$ & $\phantom{0}1{,}761$ & $\phantom{0}4{,}500$ \\
        RodoSol-ALPR~\cite{laroca2022cross} & General & $\phantom{0}9{,}989$ & $10{,}011$ & $19{,}652$ & $\phantom{0{,}}348$ & $\phantom{0}8{,}451$ & $11{,}549$ & $20{,}000$ \\
        \midrule
        Total Per Class & -- & $26{,}272$ & $58{,}563$ & $74{,}940$ & $9{,}895$ & $51{,}361$ & $33{,}474$ & $84{,}835$ \\
        \bottomrule
    \end{tabular}
\end{table*}

\section{Experimental Setup}
\label{sec:framework}

The three \gls*{uvib} tasks are treated independently as binary classification problems. We evaluate four widely adopted image-classification architectures: EfficientNetV2-S~\cite{tan2021efficientnetv2}, ResNet-50~\cite{he2016residual}, \gls*{vit}/B-16~\cite{dosovitskiy2021image}, and YOLO11s-cls~\cite{ultralytics2025yolov11cls}. Together, they represent efficient convolutional networks, residual architectures, vision transformers, and compact deployment-oriented classifiers.

All images were resized to $224 \times 224$ pixels, ensuring identical preprocessing and compatibility with ImageNet pre-training~\cite{deng2009imagenet}. The classification heads were replaced with two-output layers, and all network parameters were fine-tuned.

We used Adam with a learning rate of $10^{-4}$ and a batch size of 32. Training lasted up to 100 epochs, with early stopping after 12 epochs without improvement in validation accuracy. No data augmentation was applied, ensuring that the results reflect the training data and evaluation splits rather than augmentation-specific gains.

For the highly imbalanced \gls*{vmmr} Suitability task, cross-entropy was replaced with Focal Loss~\cite{lin2017focal}, which reduces the contribution of easy majority-class samples and emphasizes harder cases.

\subsection{Evaluation Protocols and Cross-Domain Splits}
\label{subsec:protocols}

To evaluate classification performance and generalization, we define four protocols. These protocols distinguish mixed-domain evaluation from stricter transfer settings, a key issue in real-world vehicle monitoring under dataset shift~\cite{laroca2022first, katare2025analyzing}:

\begin{itemize}
    \item \textbf{Surveillance to General (S2G)}: models are trained and validated strictly on highway surveillance camera data~($57{,}798$ images) and tested on the remaining datasets from the general domain~($27{,}037$ images);
    \item \textbf{General to Surveillance (G2S)}: training and validation occur on the general image subsets, whereas testing is performed exclusively on images captured by highway surveillance cameras;
    \item \textbf{All-Datasets}: all datasets are pooled and partitioned with stratified train, validation, and test splits, measuring performance when the training data include samples from all acquisition domains;
    \item \textbf{Cross-Dataset Shift (CDS)}: designed to distinguish domain-level variations from dataset-specific biases by providing a mix of both acquisition scenarios (Surveillance and General) in both the training and evaluation phases. Consequently, it evaluates whether the models develop a dependency on specific dataset characteristics or achieve generalized feature representation. Models are trained on a composite source domain (UFPR-VeSV, RodoSol-ALPR, Vehicle-Rear, and UFOP; $46{,}146$ samples) and evaluated on an unseen target domain (LPLCv2, UFPR-ALPR, and SSIG-SegPlate; $38{,}689$ samples).
\end{itemize}

The cross-domain protocols (S2G and G2S) use a 60/40 training-validation split within the source domain and the full target domain for testing\major{, with no vehicle-identity overlap between domains. Only 94 shared identities were found between LPLCv2 and UFPR-VeSV, both within the Surveillance domain}. The CDS protocol uses the same 60/40 split\major{, with these 94 shared identities representing less than 0.15\% of the test set}. \major{For all three protocols, target datasets are completely withheld during training and validation.}

The All-Datasets protocol uses a stratified 60/20/20 split for training, validation, and testing. Preliminary experiments with alternative ratios showed no substantial changes in performance. \major{All training splits and scripts are publicly available with the benchmark at~\urlSupplementary.}

\subsection{Statistical Rigor and Evaluation Metrics}
\label{subsec:statistical_rigor}

Each experiment was repeated with three random seeds. The protocol split remains fixed across runs, while the seed controls data shuffling and the initialization of the newly replaced classification heads\major{. This setup provides a practical estimate of the variability introduced by stochastic training factors under each evaluation protocol}. \major{Accuracy, per-class \gls*{f1}, and Macro~\gls*{f1} are computed independently from the predictions of each run and subsequently reported as mean and standard deviation~($\mu \pm \sigma$) across the three runs. This procedure enables the comparison of both predictive performance and training stability across models.}

Accuracy summarizes correctness but can be misleading for imbalanced tasks such as \gls*{vmmr} Suitability. Per-class \gls*{f1} reveals the performance obtained for each class, while Macro~\gls*{f1} assigns equal importance to both classes and is therefore the primary metric for model comparison under class imbalance. \major{Importantly, Macro~\gls*{f1} is obtained by averaging the three run-level Macro~\gls*{f1} scores, rather than by recomputing it from the aggregated and rounded per-class \gls*{f1}~values.}

\begin{table*}[htb!]
\caption{Experimental results~(\%) across the four architectures and evaluation protocols. Metrics report overall Accuracy, Macro \gls*{f1}, and per-class \gls*{f1} averaged across three independent initializations ($\mu \pm \sigma$). Bold values indicate the best Macro \gls*{f1} within each protocol and task.
}
\label{tab:experimental_results}
\centering

\vspace{-0.5mm}

\begingroup
\setlength{\tabcolsep}{5pt}
\resizebox{\textwidth}{!}{%
\begin{tabular}{llccc@{\hspace{1.0em}}c@{\hspace{1.0em}}ccc@{\hspace{1.0em}}c@{\hspace{1.0em}}ccc}
\toprule
\multirow{2}[2]{*}{\textbf{Protocol}} &
\multirow{2}[2]{*}{\textbf{Architecture}} &
\multicolumn{3}{c}{\textbf{Orientation}} & &
\multicolumn{3}{c}{\textbf{\gls*{vmmr} Suitability}} & &
\multicolumn{3}{c}{\textbf{Color Clarity}} \\
\cmidrule(r{1.0em}){3-5}
\cmidrule(l{1.0em}r{1.0em}){7-9}
\cmidrule(l{1.0em}){11-13}
 &  &
\textbf{Accuracy} & \textbf{Macro F1} & \textbf{Front / Rear} & &
\textbf{Accuracy} & \textbf{Macro F1} & \textbf{Suitable / Unsuitable} & &
\textbf{Accuracy} & \textbf{Macro F1} & \textbf{Color / Non-Color} \\
\midrule

\multirow{4}{*}{\textbf{S2G}}
 & EfficientNetV2 & 99.2 $\pm$ 0.18 & \textbf{99.2 $\pm$ 0.16} & 99.1 $\pm$ 0.20 / 99.2 $\pm$ 0.16 & & 96.8 $\pm$ 0.61 & \textbf{69.4 $\pm$ 3.53} & 98.3 $\pm$ 0.33 / 47.5 $\pm$ 4.84 & & 67.2 $\pm$ 3.02 & \textbf{75.6 $\pm$ 0.99} & 74.2 $\pm$ 1.54 / 54.7 $\pm$ \phantom{0}6.90 \\
 & ResNet-50 & 98.9 $\pm$ 0.28 & 99.0 $\pm$ 0.25 & 98.7 $\pm$ 0.31 / 99.0 $\pm$ 0.25 & & 96.2 $\pm$ 1.03 & 67.3 $\pm$ 3.19 & 98.0 $\pm$ 0.55 / 44.8 $\pm$ 4.80 & & 69.5 $\pm$ 2.60 & 75.4 $\pm$ 0.90 & 75.1 $\pm$ 1.32 / 60.5 $\pm$ \phantom{0}5.25 \\
 & ViT/B-16 & 93.7 $\pm$ 1.93 & 94.8 $\pm$ 1.41 & 92.6 $\pm$ 2.43 / 94.5 $\pm$ 1.59 & & 96.6 $\pm$ 0.61 & 68.3 $\pm$ 3.31 & 98.2 $\pm$ 0.32 / 44.9 $\pm$ 3.60 & & 62.8 $\pm$ 0.63 & 73.7 $\pm$ 0.79 & 71.9 $\pm$ 0.26 / 44.9 $\pm$ \phantom{0}1.96 \\
 & YOLO11s-cls & 90.5 $\pm$ 0.53 & 91.6 $\pm$ 0.48 & 88.8 $\pm$ 0.67 / 91.8 $\pm$ 0.44 & & 93.1 $\pm$ 0.67 & 56.6 $\pm$ 0.46 & 96.4 $\pm$ 0.37 / 21.3 $\pm$ 0.77 & & 61.9 $\pm$ 1.49 & 68.4 $\pm$ 1.42 & 70.2 $\pm$ 0.85 / 47.2 $\pm$ \phantom{0}3.25 \\

\midrule

\multirow{4}{*}{\textbf{G2S}}
 & EfficientNetV2 & 79.2 $\pm$ 1.34 & 76.4 $\pm$ 0.75 & 69.4 $\pm$ 1.25 / 84.3 $\pm$ 1.20 & & 86.9 $\pm$ 0.32 & \textbf{75.9 $\pm$ 0.65} & 92.3 $\pm$ 0.20 / 56.7 $\pm$ 1.86 & & 76.3 $\pm$ 1.48 & 74.1 $\pm$ 1.45 & 81.3 $\pm$ 1.42 / 67.7 $\pm$ \phantom{0}1.53 \\
 & ResNet-50 & 83.7 $\pm$ 3.91 & \textbf{79.7 $\pm$ 3.08} & 74.2 $\pm$ 4.40 / 88.0 $\pm$ 3.23 & & 83.2 $\pm$ 1.74 & 69.5 $\pm$ 2.97 & 89.9 $\pm$ 1.05 / 50.8 $\pm$ 5.24 & & 77.9 $\pm$ 1.46 & 75.7 $\pm$ 1.88 & 82.6 $\pm$ 0.71 / 69.6 $\pm$ \phantom{0}3.33 \\
 & ViT/B-16 & 73.1 $\pm$ 2.23 & 72.3 $\pm$ 1.11 & 62.9 $\pm$ 1.86 / 78.9 $\pm$ 2.13 & & 82.9 $\pm$ 1.21 & 65.3 $\pm$ 4.24 & 90.3 $\pm$ 0.83 / 24.2 $\pm$ 4.69 & & 75.9 $\pm$ 1.62 & 73.6 $\pm$ 2.37 & 82.3 $\pm$ 1.54 / 62.0 $\pm$ \phantom{0}1.45 \\
 & YOLO11s-cls & 83.5 $\pm$ 1.76 & 77.8 $\pm$ 2.01 & 69.5 $\pm$ 1.69 / 88.7 $\pm$ 1.41 & & 85.2 $\pm$ 0.13 & 73.3 $\pm$ 0.57 & 91.7 $\pm$ 0.07 / 33.6 $\pm$ 0.72 & & 79.7 $\pm$ 0.36 & \textbf{77.5 $\pm$ 0.39} & 84.5 $\pm$ 0.44 / 70.9 $\pm$ \phantom{0}0.25 \\

\midrule

\multirow{4}{*}{\textbf{All}}
 & EfficientNetV2 & 99.8 $\pm$ 0.01 & \textbf{99.8 $\pm$ 0.03} & 99.6 $\pm$ 0.03 / 99.8 $\pm$ 0.01 & & 96.6 $\pm$ 0.11 & \textbf{92.9 $\pm$ 0.83} & 98.1 $\pm$ 0.06 / 84.9 $\pm$ 0.58 & & 93.5 $\pm$ 0.07 & \textbf{93.2 $\pm$ 0.11} & 94.6 $\pm$ 0.05 / 91.8 $\pm$ \phantom{0}0.15 \\
 & ResNet-50 & 99.7 $\pm$ 0.04 & 99.7 $\pm$ 0.04 & 99.6 $\pm$ 0.07 / 99.8 $\pm$ 0.03 & & 96.4 $\pm$ 0.09 & 92.2 $\pm$ 0.44 & 98.0 $\pm$ 0.05 / 84.1 $\pm$ 0.57 & & 93.2 $\pm$ 0.10 & 93.0 $\pm$ 0.07 & 94.4 $\pm$ 0.07 / 91.4 $\pm$ \phantom{0}0.14 \\
 & ViT/B-16 & 99.4 $\pm$ 0.03 & 99.4 $\pm$ 0.02 & 99.1 $\pm$ 0.05 / 99.6 $\pm$ 0.02 & & 95.8 $\pm$ 0.15 & 91.2 $\pm$ 0.60 & 97.6 $\pm$ 0.08 / 81.0 $\pm$ 0.76 & & 92.6 $\pm$ 0.05 & 92.7 $\pm$ 0.12 & 94.0 $\pm$ 0.00 / 90.5 $\pm$ \phantom{0}0.16 \\
 & YOLO11s-cls & 97.2 $\pm$ 0.02 & 96.9 $\pm$ 0.07 & 95.4 $\pm$ 0.05 / 98.0 $\pm$ 0.02 & & 92.4 $\pm$ 0.05 & 84.3 $\pm$ 0.53 & 95.8 $\pm$ 0.02 / 61.6 $\pm$ 1.26 & & 90.4 $\pm$ 0.05 & 90.1 $\pm$ 0.07 & 92.1 $\pm$ 0.04 / 87.7 $\pm$ \phantom{0}0.08 \\

\midrule

\multirow{4}{*}{\textbf{CDS}}
 & EfficientNetV2 & 98.8 $\pm$ 0.26 & \textbf{96.3 $\pm$ 0.93} & 95.4 $\pm$ 0.94 / 99.3 $\pm$ 0.15 & & 89.3 $\pm$ 0.56 & \textbf{87.8 $\pm$ 2.22} & 93.6 $\pm$ 0.32 / 68.8 $\pm$ 2.76 & & 83.1 $\pm$ 0.49 & \textbf{82.2 $\pm$ 0.55} & 85.9 $\pm$ 0.56 / 78.8 $\pm$ \phantom{0}0.29 \\
 & ResNet-50 & 97.6 $\pm$ 0.44 & 92.9 $\pm$ 1.79 & 91.0 $\pm$ 1.39 / 98.6 $\pm$ 0.26 & & 85.3 $\pm$ 2.29 & 82.1 $\pm$ 6.15 & 91.0 $\pm$ 1.82 / 58.1 $\pm$ 4.97 & & 82.9 $\pm$ 1.02 & 82.1 $\pm$ 1.07 & 85.8 $\pm$ 1.08 / 78.7 $\pm$ \phantom{0}0.80 \\
 & ViT/B-16 & 95.5 $\pm$ 1.06 & 87.9 $\pm$ 2.50 & 83.8 $\pm$ 3.43 / 97.4 $\pm$ 0.63 & & 82.0 $\pm$ 1.21 & 77.7 $\pm$ 4.37 & 89.7 $\pm$ 0.64 / 28.7 $\pm$ 7.44 & & 82.5 $\pm$ 1.21 & 81.8 $\pm$ 1.41 & 85.7 $\pm$ 1.25 / 77.6 $\pm$ \phantom{0}1.02 \\
 & YOLO11s-cls & 87.4 $\pm$ 0.51 & 74.2 $\pm$ 0.53 & 64.2 $\pm$ 0.81 / 92.3 $\pm$ 0.34 & & 81.5 $\pm$ 0.13 & 73.8 $\pm$ 1.50 & 89.3 $\pm$ 0.15 / 31.7 $\pm$ 2.53 & & 81.7 $\pm$ 0.09 & 81.1 $\pm$ 0.14 & 85.3 $\pm$ 0.11 / 75.7 $\pm$ \phantom{0}0.06 \\

\bottomrule
\end{tabular}%
}
\endgroup
\end{table*}

\section{Results and Discussion}
\label{sec:results}

\cref{tab:experimental_results} compares the four architectures across the proposed protocols. The main trend is consistent: the protocol has a stronger effect on performance than the architecture. Models trained and tested with mixed-domain data perform well, whereas cross-domain evaluation exposes substantial degradation, especially for \gls*{vmmr} Suitability and Color Clarity.

The drop cannot be explained by training-set size alone. In S2G, for example, models are trained on the larger Surveillance Domain, yet performance declines on General-domain imagery. This suggests that models learn acquisition-specific cues \major{and operational conditions} that do not transfer reliably to unseen cameras, illumination patterns, and viewpoint distributions\major{, contrasting fixed traffic-camera imagery in the Surveillance Domain with the more heterogeneous viewpoints and acquisition setups of the General Domain}.

The CDS results consistently occupy an intermediate performance tier, outperforming the severe bottlenecks of one-way domain drops
while underperforming the All-Datasets baseline. Because the CDS protocol includes a balanced mixture of both Surveillance and General acquisition settings in both training and testing phases, the observed drops in Macro \gls*{f1}~(EfficientNetV2 dropping to $96.3\%$, $87.8\%$, and $82.2\%$ across the three tasks) cannot be attributed to a change in acquisition domain. Instead, these drops empirically demonstrate a strict dependency on dataset-specific biases. 

This effect is minimized in the All-Datasets setting, where joint optimization across all datasets smooths out camera-specific variances by providing a dense, continuous representation of the acquisition space.
\major{When aggregating the performance across all tasks and architectures, the cross-domain protocols exhibit a substantial increase in variance, yielding mean standard deviations of $\pm 1.4$ for S2G, $\pm 1.78$ for G2S, and $\pm 1.93$ for CDS in their Macro \gls*{f1}
metrics. This contrasts sharply with the high consistency observed in the All-Datasets protocol, which maintains a significantly lower mean standard deviation of~$0.35$. }
This \major{acquisition bias} is most visible for \gls*{vmmr} Suitability and Color Clarity, whose cues depend on occlusion severity, sensor response, lighting, and image quality. 
In contrast, mixed-domain training smooths these variations by exposing the models to a broader range of acquisition~conditions.

\begin{figure}[htb!]
\centering
    \resizebox{\linewidth}{!}{
    \;
    \begin{minipage}{0.31\linewidth}
        \centering
        \includegraphics[height=10ex]{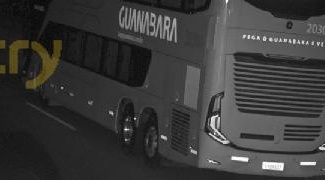} \\
        \vspace{1mm}
        \includegraphics[height=10ex]{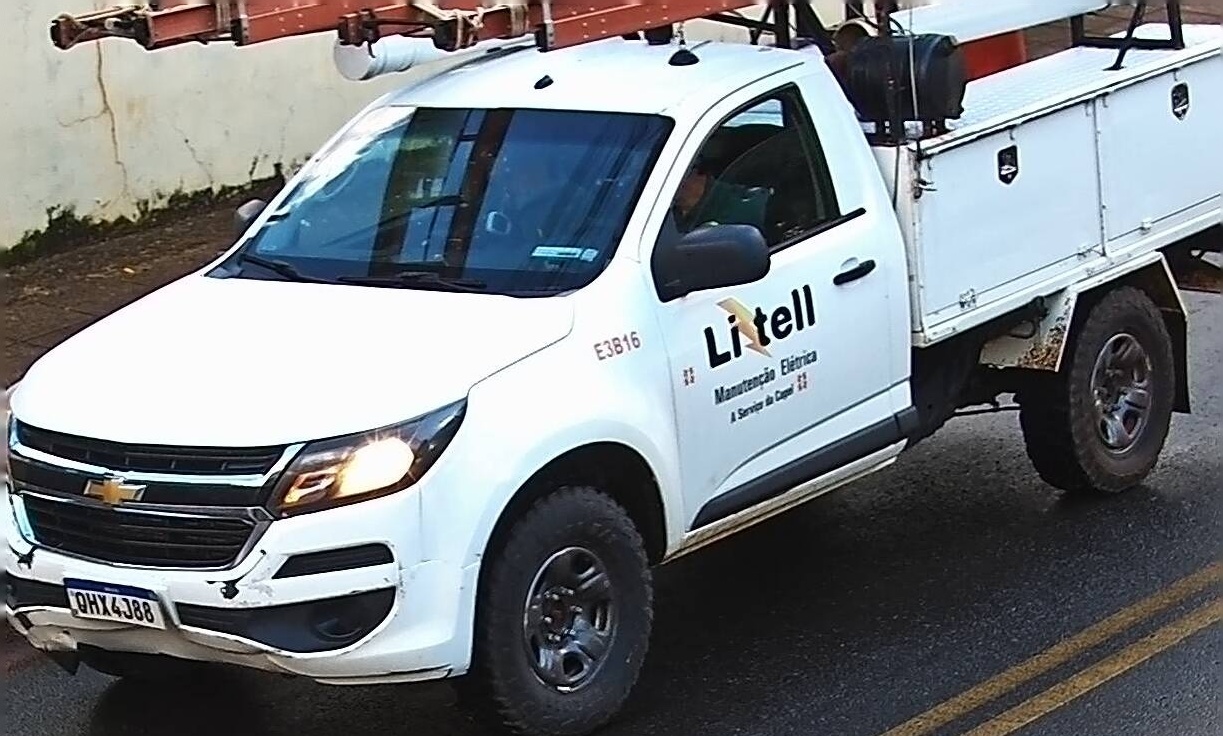} \\
        \vspace{-0.5mm}
        {\footnotesize (a) Orientation}
    \end{minipage}
    \;\;
    \begin{minipage}{0.31\linewidth}
        \centering
        \includegraphics[height=10ex]{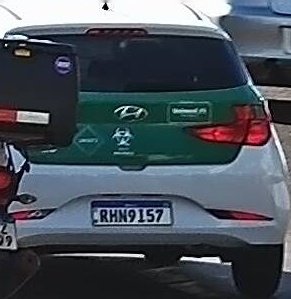} \\
        \vspace{1mm}
        \includegraphics[height=10ex]{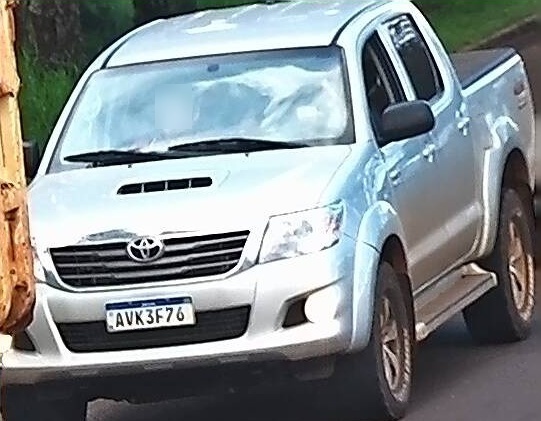} \\
        \vspace{-0.5mm}
        {\footnotesize (b) \gls*{vmmr} Suitability}
    \end{minipage}
    \begin{minipage}{0.31\linewidth}
        \centering
        \includegraphics[height=10ex]{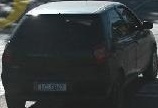} \\
        \vspace{1mm}
        \includegraphics[height=10ex]{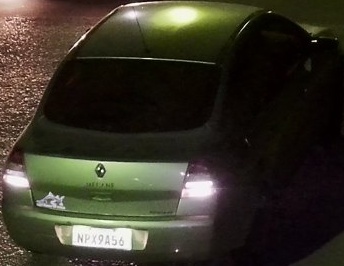} \\
        \vspace{-0.5mm}
        {\footnotesize (c) Color Clarity}
    \end{minipage}
    }

\caption{Qualitative examples of classification errors: (a)~\textit{Orientation} errors in borderline near-lateral views; (b)~\textit{\gls*{vmmr} Suitability} false positives, where partially occluded vehicles are predicted as suitable; and (c)~\textit{Color Clarity} errors in images with distorted chromatic evidence, including a green vehicle affected by glare and a silver vehicle under night~lighting.}
\label{fig:experiment_errors}
\end{figure}

\textit{Orientation} is the most transferable task. Most models achieve high Accuracy and Macro \gls*{f1}, although the Front class remains slightly weaker because rear-view vehicles are more frequent in the benchmark.
Qualitative errors are concentrated in near-lateral viewpoints, where the boundary between front and rear becomes ambiguous~(see \cref{fig:experiment_errors}a).

\textit{\gls*{vmmr} Suitability} is the most imbalanced task, and the gap between Accuracy and Macro \gls*{f1} is substantial. Even with Focal Loss, models often fail to recover the minority Unsuitable class under domain shift. In the cross-domain protocols, Macro \gls*{f1} ranges from $56.6\%$ to $75.9\%$, while Accuracy remains between $82.9\%$ and $96.8\%$. The most extreme case is 
YOLO11s-cls in S2G, whose Unsuitable \gls*{f1} drops to $21.3\%$.
This shows that high Accuracy can be dominated by the Suitable class. \major{For context, a classifier predicting only the majority class achieves an Accuracy of $88.3\%$, but a Macro \gls*{f1} of only $46.9\%$ (with $0.0\%$ for Unsuitable)}. The qualitative examples in \cref{fig:experiment_errors}b suggest that partial occlusions are often not treated as severe enough to invalidate \gls*{vmmr}~analysis.

\textit{Color Clarity} is 
sensitive to acquisition bias, but more balanced overall than \gls*{vmmr} Suitability. Its definition depends directly on exposure, sensor technology, and illumination, and most errors occur in borderline cases where chromatic information is degraded but not completely absent. Strong reflections, shadows, night lighting, and infrared-like captures can make an image unreliable for color recognition even when the vehicle is clearly visible~(\cref{fig:experiment_errors}c).

No architecture dominates all tasks and protocols, but the results offer practical guidance. EfficientNetV2-S is the strongest default when a single model is required, achieving the best or tied-best Macro \gls*{f1} in most settings while remaining highly competitive for Orientation and Color Clarity. ResNet-50 is a similarly robust alternative, particularly under All-Datasets and CDS. YOLO11s-cls is attractive when computational cost is the main constraint, achieving the best Color Clarity results under G2S, but consistently sacrificing Macro \gls*{f1} elsewhere, especially for the minority Unsuitable class. Overall, these findings reinforce that robust cross-domain evaluation matters more than selecting a stronger in-domain architecture~alone.

\section{Conclusions}
\label{sec:conclusion}

\glsreset{uvib}

This paper introduced \gls*{uvib}, a benchmark comprising \numImages images from seven public Brazilian datasets, along with new annotations for three operational vehicle-analysis tasks: Orientation, \gls*{vmmr} Suitability, and Color Clarity.
By combining surveillance and general-domain imagery, \gls*{uvib} provides public annotations, standardized baselines, and protocols for studying dataset bias and cross-domain robustness in~\gls*{its}.

The experiments show that domain shift is the main obstacle to reliable vehicle attribute recognition.
Across tasks and architectures, performance degradation is driven more by acquisition differences between training and test data than by the model family itself. While mixed-domain training yields strong results, cross-domain testing reveals persistent limitations, especially for \gls*{vmmr} Suitability and Color Clarity, with the cross-dataset protocol being an intermediate alternative.
\gls*{vmmr} Suitability is affected by severe class imbalance and ambiguous partial occlusions, whereas Color Clarity depends strongly on illumination, exposure, and sensor modality. These findings show why Accuracy alone is insufficient and why Macro \gls*{f1} should be considered when evaluating operational vehicle-analysis tasks.

The results also suggest practical starting points for future systems. EfficientNetV2-S is a strong default baseline, ResNet-50 remains highly competitive, and compact classifiers such as YOLO11s-cls should be selected mainly when efficiency is prioritized over minority-class robustness. More broadly, progress will depend not only on improved architectures, but also on methods that mitigate dataset bias and generalize across heterogeneous acquisition conditions.

Future work can extend \gls*{uvib} in three main directions. First, the annotation framework can be refined by introducing multiple occlusion-severity levels and richer color-quality categories. Second, these operational tasks can be incorporated into downstream \gls*{vmmr} and \gls*{vcr} pipelines to assess whether filtering or joint prediction improves end-task reliability. Third, multi-task learning, synthetic data generation, and targeted augmentation strategies can be explored to improve rare-condition coverage and reduce cross-domain~degradation.

\section*{\uppercase{Acknowledgments}}

\iffinal
    This study was financed in part by the \textit{Coordenação de Aperfeiçoamento de Pessoal de Nível Superior - Brasil~(CAPES)}, through the \textit{Programa de Excelência Acadêmica~(PROEX)} - Finance Code 001, in part by the \textit{Fundação Araucária} under grant \#~078/2026, and in part by the \textit{Conselho Nacional de Desenvolvimento Científico e Tecnológico~(CNPq)} (\#~315409/2023-1).
\fi

\bibliographystyle{IEEEtran}
\bibliography{bibtex}

\end{document}